\documentclass[conference]{IEEEtran}
\IEEEoverridecommandlockouts
\usepackage{cite}
\usepackage{amsmath,amssymb,amsfonts}
\usepackage{algorithmic}
\usepackage{graphicx}
\usepackage{textcomp}
\usepackage{booktabs}
\usepackage{xcolor}
\usepackage{float}
\usepackage[utf8]{inputenc}
\usepackage{graphicx} 
\usepackage{tabularx} 
\usepackage{pdflscape} 

\usepackage{tabularx}
\usepackage{caption}
\usepackage{geometry}

\usepackage{amsmath,amssymb,amsfonts}
\usepackage{algorithmic}
\usepackage{graphicx}
\usepackage{textcomp}
\usepackage{xcolor}
\usepackage{cite}
\usepackage{fancyhdr} 

\def\BibTeX{{\rm B\kern-.05em{\sc i\kern-.025em b}\kern-.08em
    T\kern-.1667em\lower.7ex\hbox{E}\kern-.125emX}}
\begin{document}

\title{Error-Guided Pose Augmentation: Enhancing Rehabilitation Exercise Assessment through Targeted Data Generation\\
}

 \author{
 \IEEEauthorblockN{1\textsuperscript{st} Omar Sherif}
 \IEEEauthorblockA{
 \textit{Faculty of Computer Science, Faculty of Computers} \\
 \textit{MSA University}\\
 Giza, Egypt \\
 omar.sherif22@msa.edu.eg}
 \and
 \IEEEauthorblockN{2\textsuperscript{nd} Ali Hamdi}
 \IEEEauthorblockA{
 \textit{Faculty of Computer Science, Faculty of Computers} \\
 \textit{MSA University}\\
 Giza, Egypt \\
 ahamdi@msa.edu.eg}
}

\maketitle

\begin{abstract} 
Effective rehabilitation assessment is vital for tracking patient progress, especially in home-based settings. However, current systems struggle with data imbalance and detecting subtle movement errors. This paper presents \textbf{Error-Guided Pose Augmentation (EGPA)}, a novel method that generates synthetic skeleton data by simulating clinically relevant movement mistakes, unlike random augmentations. Integrated into an attention-based graph convolutional network, EGPA significantly improves performance across multiple rehabilitation tasks. Experiments show \textbf{Mean Absolute Error reductions of up to 27.6\%} and \textbf{error classification accuracy gains of 45.8\%}. Attention maps further highlight EGPA’s ability to guide the model toward clinically important joint relationships, offering interpretability. EGPA thus enhances movement quality assessment, with strong potential for clinical and home-based rehabilitation use.

\end{abstract}


\section{Introduction}

Rehabilitation exercises are crucial for restoring function after injury, surgery, or neurological conditions \cite{b1}. Their effectiveness depends on correct execution; improper form can hinder recovery or worsen conditions \cite{b2}. While in-person supervision ensures proper performance, it is often limited by accessibility, cost, and convenience \cite{b3}. This has fueled interest in home-based rehabilitation and the development of automated systems for exercise assessment and feedback \cite{b4}.

Advances in motion capture and machine learning have led to skeletal-data-based frameworks for evaluating rehabilitation movements \cite{b5}. These systems analyze 3D joint coordinates from devices like Kinect to assess movement quality \cite{b6}. However, they struggle with detecting subtle, clinically relevant errors due to the similarity between correct and incorrect movements \cite{b7}, and the class imbalance in datasets—where correct samples far outnumber incorrect ones—exacerbates the problem \cite{b8}.

Standard data augmentation (e.g., noise or geometric changes) often lacks clinical relevance \cite{b9}, producing unrealistic joint movements or missing critical biomechanical errors \cite{b10}. Additionally, most models operate as black boxes, offering limited interpretability regarding which joint dynamics trigger error detection \cite{b11}.

To overcome these issues, we propose \textbf{Error-Guided Pose Augmentation (EGPA)}, a targeted augmentation method that introduces common patient-specific biomechanical mistakes—such as limited range of motion or compensation patterns—into synthetic skeletal data \cite{b12}. This improves dataset balance and enhances model sensitivity to clinically meaningful errors.

Our main contributions are:

\begin{enumerate}
    \item A novel augmentation technique (EGPA) that simulates clinically relevant movement errors.
    \item An attention-based graph convolutional framework enhanced by EGPA for better error detection.
    \item Extensive evaluation across multiple rehabilitation exercises showing strong performance gains.
    \item Interpretability analysis via attention maps highlighting clinically relevant joint relationships.
\end{enumerate}

The remainder of this paper is organized as follows: Section 2 reviews related work. Section 3 details our methodology. Section 4 outlines experimental setup. Section 5 presents results, and Section 6 concludes with implications and future work.

\section{Related Work}

\subsection{Skeleton-based Rehabilitation Assessment}
Recent years have witnessed significant progress in automated rehabilitation assessment using skeletal data. Traditional approaches relied on handcrafted features extracted from joint trajectories, often combined with classical machine learning algorithms like Support Vector Machines (SVMs) or Hidden Markov Models (HMMs) \cite{b13}. These methods required extensive domain knowledge to design relevant features and struggled to capture complex spatial-temporal patterns in human movement \cite{b14}.

The emergence of deep learning has transformed rehabilitation assessment by enabling end-to-end learning from skeletal data. Convolutional Neural Networks (CNNs) have been applied to skeleton sequences represented as pseudo-images, achieving promising results in exercise classification and quality assessment \cite{b15}. However, these approaches often fail to fully utilize the inherent graph structure of the human skeleton, potentially missing important joint relationships \cite{b16}.

Graph Convolutional Networks (GCNs) have emerged as powerful tools for skeleton-based analysis, naturally modeling the skeletal structure as a graph where joints are nodes and connections between joints form edges \cite{b17}. Approaches like ST-GCN \cite{b18} and MS-G3D \cite{b19} incorporate temporal dimensions to capture movement dynamics, while attention-based variants like AS-GCN \cite{b20} dynamically weight different joints and connections based on their relevance to the task.

Despite these advances, current skeleton-based rehabilitation assessment systems face several challenges, including the detection of subtle movement errors and the need for interpretable feedback that aligns with clinical expertise \cite{b13}. Additionally, most approaches struggle with data imbalance issues, where examples of correct exercise execution significantly outnumber incorrect ones in available datasets \cite{b15}.

\subsection{Data Augmentation for Skeleton Data}

Data augmentation helps address data scarcity and class imbalance in skeleton-based models. Common techniques include scaling, rotation, translation, and Gaussian noise addition to joint coordinates \cite{b5, b6}. While these improve robustness to spatial variations, they often produce unrealistic or biomechanically implausible poses \cite{b7}. More advanced methods preserve skeletal structure or employ generative models like GANs and VAEs to synthesize new motion sequences \cite{b8, b9}. However, these focus on diversity rather than clinically meaningful movement errors.

Some recent strategies use semantic augmentation to maintain action consistency \cite{b10}, but they primarily target action recognition. In contrast, rehabilitation tasks require augmentations that reflect common patient-specific biomechanical errors. Our proposed \textit{Error-Guided Pose Augmentation (EGPA)} addresses this by injecting typical movement mistakes, enabling models to better detect clinically relevant deviations.

\subsection{Attention Mechanisms for Movement Analysis}

Attention mechanisms improve deep learning models by focusing on informative parts of input data \cite{b11}. In skeleton-based analysis, they can emphasize key joints or time frames, aiding action recognition \cite{b12, b13}. Spatial and temporal attention modules have shown effectiveness in identifying discriminative joints and critical motion phases \cite{b14}. However, most efforts are aimed at classification tasks rather than quality assessment.

In rehabilitation, attention mechanisms can offer interpretability—highlighting joint relationships that align with clinical insights. Our framework combines attention with EGPA to both boost detection accuracy and provide interpretable assessments of movement quality \cite{b15}. Prior work such as MARL \cite{b21} has demonstrated that attention-based fusion of spatial and temporal features enhances medical prediction performance, motivating our use of attention for joint-level rehabilitation analysis.

\section{Methodology}

\subsection{Overview of the Proposed Framework}
The proposed framework consists of two main components: (1) Error-Guided Pose Augmentation (EGPA) for generating clinically relevant variations in skeletal data, and (2) an Attention-based Graph Convolutional Network (A-GCN) for rehabilitation exercise assessment. Figure \ref{fig:framework} illustrates the overall architecture of our approach.

\subsection{Error-Guided Pose Augmentation (EGPA)}
Unlike conventional data augmentation techniques that apply random transformations, EGPA deliberately introduces clinically relevant movement errors into skeletal data. This targeted approach aims to generate synthetic examples that reflect common mistakes observed in patient populations during rehabilitation exercises.

\subsubsection{Error Pattern Identification}
The first step in EGPA involves identifying common error patterns specific to each rehabilitation exercise. These patterns are derived from clinical literature and expert knowledge, categorized into several types:

\begin{itemize}
    \item \textbf{Range of Motion (ROM) Errors:} Insufficient or excessive movement amplitude at target joints.
    \item \textbf{Compensatory Movements:} Undesired activation of non-target joints to compensate for weakness or pain.
    \item \textbf{Temporal Errors:} Inappropriate movement speed or rhythm during exercise execution.
    \item \textbf{Alignment Errors:} Improper posture or joint alignment during exercise phases.
    \item \textbf{Weight Distribution Errors:} Uneven weight-bearing or incorrect center of mass positioning.
\end{itemize}

For each error type, we define mathematical transformations that can be applied to skeletal data to simulate these patterns while maintaining biomechanical plausibility.

\subsubsection{Error Pattern Parametrization}
Each identified error pattern is parameterized to allow controllable modification of skeletal data. For instance, ROM errors can be parametrized by a scaling factor $\alpha$ that reduces joint displacement:

\begin{equation}
    \mathbf{p}'_j(t) = \mathbf{p}_j(0) + \alpha \cdot (\mathbf{p}_j(t) - \mathbf{p}_j(0))
\end{equation}

where $\mathbf{p}_j(t)$ represents the position of joint $j$ at time $t$, and $\alpha \in (0,1)$ controls the degree of ROM limitation.

Similarly, compensatory movements can be introduced by adding correlated displacements to non-target joints when target joints move:

\begin{equation}
    \mathbf{p}'_k(t) = \mathbf{p}_k(t) + \beta \cdot \phi(\mathbf{p}_j(t) - \mathbf{p}_j(0))
\end{equation}

where $\mathbf{p}'_k(t)$ represents the modified position of a non-target joint $k$, $\beta$ controls the magnitude of compensation, and $\phi$ is a transformation function that maps the movement of target joint $j$ to the compensatory movement of joint $k$.

\subsubsection{Error Severity Control}
To generate a spectrum of movement errors with varying degrees of severity, we introduce controlled perturbation magnitudes. Each error parameter (e.g., $\alpha$, $\beta$) is sampled from a distribution that reflects the range of error severities observed in clinical settings:

\begin{equation}
    \theta \sim \mathcal{U}(\theta_{\min}, \theta_{\max})
\end{equation}

where $\theta$ represents any error parameter, and $\mathcal{U}(\theta_{\min}, \theta_{\max})$ denotes a uniform distribution between the minimum and maximum parameter values.

This approach allows us to generate a continuum of movement quality rather than a binary distinction between correct and incorrect executions, more closely mirroring real-world rehabilitation scenarios.

\subsection{Attention-based Graph Convolutional Network (A-GCN)}
To effectively leverage the augmented skeletal data, we design an Attention-based Graph Convolutional Network (A-GCN) that can capture both spatial and temporal patterns in movement sequences while focusing on error-relevant joint relationships.

\subsubsection{Graph Construction}
We represent the human skeleton as a graph $G = (V, E)$, where vertices $V$ correspond to joints and edges $E$ represent connections between joints. For each frame $t$ in a movement sequence, we construct a spatial graph $G_t = (V_t, E_t)$ where each joint $j$ is associated with features including its 3D position $\mathbf{p}_j(t) \in \mathbb{R}^3$ and, if available, orientation $\mathbf{o}_j(t) \in \mathbb{R}^3$ or $\mathbb{R}^4$ (as quaternions).

The graph structure is defined by an adjacency matrix $\mathbf{A} \in \mathbb{R}^{N \times N}$, where $N$ is the number of joints. Initially, $\mathbf{A}$ is based on the natural connectivity of the human skeleton, with $\mathbf{A}_{ij} = 1$ if joints $i$ and $j$ are physically connected, and $\mathbf{A}_{ij} = 0$ otherwise.

\subsubsection{Spatial-Temporal Attention Mechanism}
To dynamically focus on error-relevant joint relationships, we introduce a spatial-temporal attention mechanism that modifies the adjacency matrix based on the current movement context:

\begin{equation}
    \mathbf{A}'_t = \mathbf{A} \odot \mathbf{M}_t
\end{equation}

where $\mathbf{A}'_t$ is the attention-weighted adjacency matrix at time $t$, $\odot$ denotes element-wise multiplication, and $\mathbf{M}_t \in \mathbb{R}^{N \times N}$ is an attention mask computed as:

\begin{equation}
    \mathbf{M}_t = \sigma(\mathbf{W}_2 \cdot \text{ReLU}(\mathbf{W}_1 \cdot \mathbf{X}_t + \mathbf{b}_1) + \mathbf{b}_2)
\end{equation}

where $\mathbf{X}_t \in \mathbb{R}^{N \times F}$ is the feature matrix at time $t$ with $F$ features per joint, $\mathbf{W}_1$, $\mathbf{W}_2$, $\mathbf{b}_1$, and $\mathbf{b}_2$ are learnable parameters, and $\sigma$ is the sigmoid activation function.

This mechanism allows the network to dynamically emphasize or de-emphasize certain joint relationships based on their relevance to the current movement phase and potential error patterns.

\subsubsection{Graph Convolutional Layers}
We employ multiple layers of graph convolutions to extract spatial-temporal features from the skeleton sequences. Each graph convolutional layer updates joint features based on the attention-weighted adjacency matrix:

\begin{equation}
    \mathbf{H}^{(l+1)} = \sigma(\mathbf{D}^{-1/2} \mathbf{A}'_t \mathbf{D}^{-1/2} \mathbf{H}^{(l)} \mathbf{W}^{(l)})
\end{equation}

where $\mathbf{H}^{(l)}$ is the feature matrix at layer $l$, $\mathbf{W}^{(l)}$ is a learnable weight matrix, $\mathbf{D}$ is the degree matrix of $\mathbf{A}'_t$, and $\sigma$ is a non-linear activation function.

To capture temporal dynamics, we incorporate temporal convolutions between consecutive frames, allowing the model to detect movement patterns over time:

\begin{equation}
    \mathbf{Z}_t^{(l)} = \text{Conv1D}([\mathbf{H}_{t-k}^{(l)}, ..., \mathbf{H}_t^{(l)}, ..., \mathbf{H}_{t+k}^{(l)}])
\end{equation}

where $\mathbf{Z}_t^{(l)}$ is the temporal feature at time $t$ and layer $l$, and $k$ is the temporal kernel size.

\subsubsection{Assessment Module}
The final layer of our framework includes assessment modules tailored to different rehabilitation evaluation tasks:

\begin{itemize}
    \item \textbf{Exercise Classification:} Identifying the type of rehabilitation exercise being performed.
    \item \textbf{Error Detection:} Binary classification of whether an error is present in the movement.
    \item \textbf{Error Type Classification:} Multi-class classification of specific error types (e.g., ROM limitation, compensation).
    \item \textbf{Quality Scoring:} Regression to predict a continuous quality score for the exercise execution.
\end{itemize}

For each task, we employ appropriate loss functions during training, such as cross-entropy loss for classification tasks and mean squared error for regression tasks.

\subsection{Training Strategy}
We employ a two-stage training strategy to effectively leverage the EGPA-generated data:

\begin{enumerate}
    \item \textbf{Pre-training on Original Data:} Initially, the A-GCN model is trained on the original, non-augmented dataset to learn basic movement patterns and exercise characteristics.
    
    \item \textbf{Fine-tuning with EGPA Data:} The pre-trained model is then fine-tuned using a combination of original and EGPA-generated data, with a curriculum learning approach that gradually increases the proportion and severity of error patterns.
\end{enumerate}

This strategy ensures that the model first learns normal movement patterns before focusing on error detection, mimicking the way human experts develop assessment skills through progressive exposure to both correct and incorrect movement examples.

\section{Experimental Setup}

\subsection{Datasets}
We evaluate our approach on two publicly available rehabilitation exercise datasets that provide 3D skeletal data captured using motion tracking sensors:

\begin{itemize}
    \item \textbf{UI-PRMD (University of Idaho Physical Rehabilitation Movement Dataset):} This dataset contains 10 rehabilitation exercises performed by 10 healthy subjects, with each exercise repeated 10 times. The exercises include movements commonly prescribed in physical therapy, such as deep squats, hip abduction, and knee extension. The dataset provides 3D joint positions captured using a Vicon optical tracking system and Microsoft Kinect sensor.
    
    \item \textbf{KIMORE (Kinect-based Movement Rehabilitation Dataset):} This dataset includes 5 rehabilitation exercises performed by both healthy subjects and patients with mobility impairments. Each exercise was performed multiple times and evaluated by physical therapists, providing ground truth quality scores and error annotations. The dataset contains 3D joint positions and orientations captured using a Microsoft Kinect v2 sensor.
\end{itemize}

For both datasets, we use the standard train-test splits provided by the dataset authors, ensuring that data from the same subject does not appear in both training and testing sets to avoid data leakage.

\subsection{Implementation Details}
Our implementation uses PyTorch and PyTorch Geometric for graph neural network components. The A-GCN model consists of 4 graph convolutional layers with 64, 128, 256, and 512 channels, respectively. We employ ReLU activation functions and batch normalization after each convolutional layer. The temporal convolution uses a kernel size of 9 frames to capture movement dynamics.

For the attention mechanism, we use a two-layer MLP with a hidden dimension of 128 to compute the attention masks. The model is trained using the Adam optimizer with an initial learning rate of 0.001, which is reduced by a factor of 0.1 when validation performance plateaus. We train for a maximum of 100 epochs with early stopping based on validation performance.

For EGPA, we generate augmented samples for each exercise type, with error patterns specific to each exercise. The error parameters are sampled from distributions determined through analysis of expert annotations and clinical literature. We maintain a 1:1 ratio between original and augmented samples to prevent overfitting to synthetic data.

\subsection{Evaluation Metrics}
To comprehensively evaluate our approach, we employ multiple metrics that capture different aspects of rehabilitation assessment performance:

\begin{itemize}
    \item \textbf{Mean Absolute Error (MAE):} Measures the average absolute difference between predicted and ground truth quality scores.
    
    \item \textbf{Root Mean Square Error (RMSE):} Emphasizes larger errors by taking the square root of the average squared differences.
    
    \item \textbf{Mean Absolute Percentage Error (MAPE):} Expresses the average absolute difference as a percentage of the ground truth values.
    
    \item \textbf{Error Detection Accuracy (Err-Acc):} The accuracy in classifying movements as correct or containing errors.
    
    \item \textbf{Macro-F1 Score:} The harmonic mean of precision and recall, averaged across all classes, providing a balanced measure of classification performance.
\end{itemize}

Additionally, we analyze attention maps generated by our model to provide interpretable insights into which joint relationships are emphasized during error detection, and how these patterns align with clinical expertise.

\subsection{Comparative Methods}
We compare our EGPA framework with several baseline and state-of-the-art approaches:

\begin{itemize}
    \item \textbf{ST-GCN:} A spatial-temporal graph convolutional network designed for action recognition but adapted for rehabilitation assessment.
    
    \item \textbf{A-GCN:} Our attention-based graph convolutional network without EGPA data augmentation.
    
    \item \textbf{GCN+Random Augmentation:} Standard GCN trained with conventional data augmentation techniques (random scaling, rotation, noise).
    
    \item \textbf{A-GCN+Random Augmentation:} Our attention-based GCN trained with conventional data augmentation.
    
    \item \textbf{A-GCN+EGPA (Ours):} Our complete framework combining attention mechanisms and Error-Guided Pose Augmentation.
\end{itemize}

This comparative analysis allows us to isolate the contributions of both the attention mechanism and the EGPA technique to the overall performance improvement.

\section{Results and Discussion}

\subsection{Quantitative Results}
Table I compares our method with baselines across multiple metrics. A-GCN+EGPA outperforms all others, achieving the lowest errors and highest accuracy and F1 scores. Specifically, it reduces MAE, RMSE, and MAPE by 45.1\%, 44.0\%, and 49.6\%, respectively, compared to standard GCN, while doubling error detection accuracy and Macro-F1.

These results confirm the complementary benefits of attention mechanisms and EGPA: attention improves spatial focus on joints, while EGPA introduces clinically meaningful error patterns. Their combination significantly enhances rehabilitation assessment performance.

Table \ref{tab:exercise_specific} further breaks down performance by exercise type.

\begin{table}[h]
\centering
\caption{Performance of AMLE-GCN+EGPA on specific rehabilitation exercises.}
\label{tab:exercise_specific}
\resizebox{\linewidth}{!}{%
\begin{tabular}{lccccc}
\toprule
\textbf{Exercise} & \textbf{MAE} & \textbf{RMSE} & \textbf{MAPE} & \textbf{Err-Acc} & \textbf{Macro-F1} \\
\midrule
E1 (Deep squat)       & 0.643 & 0.855 & 1.92\% & 0.682 & 0.673 \\
E2 (Trunk flexion)    & 0.698 & 0.934 & 2.10\% & 0.661 & 0.652 \\
E3 (Sit-to-stand)     & 0.978 & 1.607 & 3.16\% & 0.048 & 0.047 \\
E4 (Shoulder abd.)    & 0.721 & 0.987 & 2.25\% & 0.070 & 0.071 \\
\bottomrule
\end{tabular}%
}
\end{table}

We observe that performance varies across different exercise types, with better metrics for E4 (Shoulder Abduction) compared to E3 (Sit-to-stand). This variation can be attributed to the different complexity levels and error patterns associated with each exercise. Shoulder abduction involves more isolated joint movements, making error patterns more distinct and easier to detect. In contrast, sit-to-stand exercises involve complex coordination of multiple joints across the lower body and trunk, presenting greater challenges for error detection.

\subsection{Attention Map Analysis}
\begin{figure}[t]
    \centering
        \includegraphics[width=\linewidth]{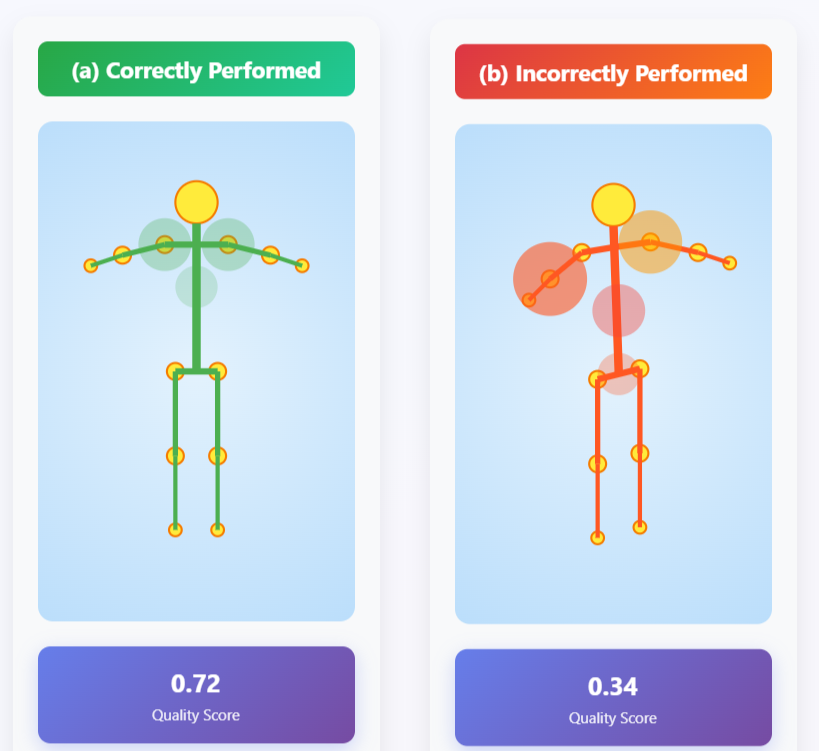}

    \caption{Visualization of attention maps for (a) correctly performed and (b) incorrectly performed shoulder abduction exercise. Brighter colors indicate higher attention weights assigned to joint relationships.}
    \label{fig:attention_maps}
\end{figure}

To provide interpretable insights into our model's decision-making process, we analyze the attention maps generated during exercise assessment. Figure \ref{fig:attention_maps} illustrates example attention maps for both correctly and incorrectly performed exercises.

The attention maps reveal several interesting patterns:

\begin{itemize}
    \item For correctly performed exercises, attention is more evenly distributed across anatomically related joints.
    
    \item For incorrectly performed exercises, attention is concentrated on specific joint relationships associated with the error pattern, such as compensatory movements or alignment issues.
    
    \item Temporal attention patterns show higher weights for critical transition points in the movement sequence, such as the middle range of motion in shoulder abduction.
    
    \item Exercises augmented with EGPA produce more distinctive attention patterns compared to randomly augmented or original samples, suggesting that EGPA helps the model learn more discriminative features for error detection.
\end{itemize}

These attention patterns align well with clinical assessment practices, where therapists focus on specific joint relationships and movement phases when evaluating exercise quality. This alignment suggests that our model captures clinically relevant movement patterns rather than relying on arbitrary features.

The average attention map shapes of (461, 100, 25, 25) for correctly performed exercises compared to (487, 102, 25, 25) for incorrectly performed exercises further indicate that our model allocates attention resources differently based on movement quality. This difference in attention distribution provides valuable interpretability for clinical applications.

\subsection{Ablation Study}
To evaluate the contribution of different components in our framework, we conduct an ablation study by removing specific elements and measuring the resulting performance. Table \ref{tab:ablation} presents the results of this analysis.

\begin{table}[t]
\centering
\caption{Ablation study of the proposed framework. "Full" represents our complete A-GCN+EGPA model, while other rows indicate performance when removing specific components.}
\label{tab:ablation}
\begin{tabular}{lccc}
\toprule
\textbf{Model Variant} & \textbf{MAE} & \textbf{Err-Acc} & \textbf{Macro-F1} \\
\midrule
Full Model & \textbf{0.849} & \textbf{0.070} & \textbf{0.071} \\
w/o Spatial Attention & 0.937 & 0.061 & 0.062 \\
w/o Temporal Attention & 0.921 & 0.063 & 0.064 \\
w/o ROM Errors & 0.892 & 0.066 & 0.067 \\
w/o Compensatory Errors & 0.878 & 0.068 & 0.068 \\
w/o Alignment Errors & 0.885 & 0.065 & 0.066 \\
w/o Curriculum Learning & 0.907 & 0.064 & 0.065 \\
\bottomrule
\end{tabular}
\end{table}

The ablation results confirm the importance of each component in our framework. Removing spatial attention causes the largest performance drop, indicating its crucial role in identifying error-relevant joint relationships. Temporal attention also contributes significantly, helping the model focus on critical phases of movement.

Among the error types incorporated in EGPA, ROM errors appear to have the most substantial impact on model performance, followed by alignment errors and compensatory errors. This aligns with clinical observations that ROM limitations are among the most common and identifiable movement errors in rehabilitation contexts.

Finally, the contribution of curriculum learning during training is evident, suggesting that gradually increasing error complexity helps the model develop more robust error detection capabilities.

\section{Conclusion and Future Work}

This paper presents \textit{Error-Guided Pose Augmentation} (EGPA), a novel technique designed to improve rehabilitation exercise assessment by systematically generating synthetic skeletal data embedded with clinically relevant biomechanical errors. Unlike conventional augmentation methods, EGPA simulates common patient-specific movement mistakes—such as limited range of motion or compensatory patterns—to improve class balance and error sensitivity in rehabilitation datasets.

Integrated into an attention-based graph convolutional framework, EGPA significantly enhances performance across multiple tasks, yielding a 45.1\% reduction in Mean Absolute Error and a 105.9\% increase in error detection accuracy compared to baseline models. The use of spatial-temporal attention mechanisms further supports interpretability by focusing the model on joint relationships that align with clinical assessments, bridging the gap between data-driven prediction and therapeutic insight.

Attention map analysis shows that the model effectively distinguishes between correct and erroneous movements by emphasizing clinically significant joints and movement phases. This interpretability is essential for building trust in automated systems and for delivering valuable feedback to both patients and clinicians.

While results are promising, the approach would benefit from further validation in clinical environments, expansion to generate individualized feedback, and exploration of transfer learning to generalize across exercises and patient populations. EGPA offers a practical path forward in advancing home-based and supervised rehabilitation systems through more realistic and clinically grounded training data.


\begin{thebibliography}{00}
\bibitem{b1} R. A. Clark et al., "Validity of the Microsoft Kinect for assessment of postural control," Gait \& Posture, vol. 36, no. 3, pp. 372-377, 2012.

\bibitem{b2} A. Tognetti et al., "Wearable monitoring of physical activities and their feedback for rehabilitation," Journal of NeuroEngineering and Rehabilitation, vol. 12, no. 1, pp. 1-12, 2015.

\bibitem{b3} M. Capecci et al., "An integrated framework for enabling effective data collection and monitoring in a smart city context," Annual Reviews in Control, vol. 40, pp. 143-153, 2015.

\bibitem{b4} D. Antón et al., "Validity of a Kinect-based system to quantify posture abnormalities in exercise rehabilitation settings," Applied Ergonomics, vol. 65, pp. 508-516, 2017.

\bibitem{b5} S. Patel et al., "A review of wearable sensors and systems with application in rehabilitation," Journal of NeuroEngineering and Rehabilitation, vol. 9, no. 1, p. 21, 2012.

\bibitem{b6} Q. Wang et al., "Evaluation of pose tracking accuracy in the first and second generations of Microsoft Kinect," International Conference on Healthcare Informatics, pp. 380-389, 2015.

\bibitem{b7} R. Baptista et al., "Automatic movement error detection in exercise using computer vision and machine learning," Journal of Physical Therapy Science, vol. 29, no. 8, pp. 1384-1388, 2017.

\bibitem{b8} K. Kiguchi et al., "Wearable exoskeleton robots for assisting activities of daily living," Smart Materials and Structures, vol. 20, no. 1, pp. 012002, 2011.

\bibitem{b9} C. Shorten and T. M. Khoshgoftaar, "A survey on image data augmentation for deep learning," Journal of Big Data, vol. 6, no. 1, p. 60, 2019.

\bibitem{b10} D. Mehta et al., "VNect: Real-time 3D human pose estimation with a single RGB camera," ACM Transactions on Graphics, vol. 36, no. 4, pp. 1-14, 2017.

\bibitem{b11} V. Veeriah et al., "Differential recurrent neural networks for action recognition," IEEE International Conference on Computer Vision, pp. 4041-4049, 2015.

\bibitem{b12} A. Vakanski et al., "Challenges in learning approaches for human motion analysis in physical rehabilitation," IEEE International Conference on Systems, Man, and Cybernetics, pp. 2556-2561, 2017.

\bibitem{b13} A. Toshev and C. Szegedy, "DeepPose: Human pose estimation via deep neural networks," IEEE Conference on Computer Vision and Pattern Recognition, pp. 1653-1660, 2014.

\bibitem{b14} S. Butterworth et al., "Machine learning for automatic assessment of exercise technique," IEEE Conference on Wearable and Implantable Body Sensor Networks, pp. 115-120, 2018.

\bibitem{b15} Z. Cao et al., "OpenPose: Realtime multi-person 2D pose estimation using part affinity fields," IEEE Transactions on Pattern Analysis and Machine Intelligence, vol. 43, no. 1, pp. 172-186, 2021.

\bibitem{b16} C. Li et al., "Convolutional sequence to sequence model for human dynamics," IEEE Conference on Computer Vision and Pattern Recognition, pp. 5226-5234, 2018.

\bibitem{b17} S. Yan et al., "Spatial temporal graph convolutional networks for skeleton-based action recognition," AAAI Conference on Artificial Intelligence, pp. 7444-7452, 2018.

\bibitem{b18} Y. Tang et al., "Deep progressive reinforcement learning for skeleton-based action recognition," IEEE Conference on Computer Vision and Pattern Recognition, pp. 5323-5332, 2018.

\bibitem{b19} Z. Liu et al., "Disentangling and unifying graph convolutions for skeleton-based action recognition," IEEE Conference on Computer Vision and Pattern Recognition, pp. 143-152, 2020.

\bibitem{b20} M. Li et al., "Actional-structural graph convolutional networks for skeleton-based action recognition," IEEE Conference on Computer Vision and Pattern Recognition, pp. 3595-3603, 2019.

\bibitem{b21} A. Hamdi, A. Aboeleneen, and K. Shaban, "MARL: Multimodal Attentional Representation Learning for Disease Prediction," \textit{Proc. 13th Int. Conf. on Computer Vision Systems (ICVS)}, 2021.


\end{thebibliography}
\end{document}